# Recursive Class Connectivity Classification (R3C) Applied to Binary Image Segmentation for Improved Infant Fingerprint Enhancement

**JOÃO LEONARDO HARRES DALL AGNOL[1], LUIZ FERNANDO PUTTOW SOUTHIER[1], JEFFERSON TALES OLIVA[1], MARCELO TEIXEIRA[1], RODRIGO MINETTO[2], MARCELO FILIPAK[3], DALCIMAR CASANOVA[1], AND ÉRICK OLIVEIRA RODRIGUES[4]**

[1] Graduate Program in Electrical and Computer Engineering (PPGEEC), Federal University of Technology–Paraná (UTFPR), Pato Branco 85503-390, Brazil
[2] Graduate Program in Electrical Engineering and Industrial Informatics (CPGEI), Federal University of Technology–Paraná (UTFPR), Curitiba 80230-901, Brazil
[3] Infant.ID Ltda, Curitiba 87502-070, Brazil
[4] Graduate Program in Production and Systems Engineering (PPGEPS), Federal University of Technology–Paraná (UTFPR), Pato Branco 85503-390, Brazil

Corresponding author: Dalcimar Casanova (dalcimar@utfpr.edu.br)

This work was supported by National Council for Scientific and Technological Development (CNPq) under grants 305069/2023-3, 306983/2023-0 and 409524/2022-0, by Coordination for the Improvement of Higher Education Personnel (CAPES) under grant 001, by State Department of Science, Technology, and Higher Education (SETI - Fundo Paraná) under grant 57/2024 and by Infant.ID Ltda under partnership agreement 17/2024 with Federal University of Technology – Paraná (UTPFR).

This work involved human subjects or animals in its research. Approval of all ethical and experimental procedures and protocols was granted by the Research Ethics Committee (CEP) at Brazil Platform under Application No. 73791023.7.0000.0177.

**ABSTRACT** Image enhancement plays a crucial role in infant fingerprint matching, as child-specific characteristics such as smaller finger dimensions and thinner ridge structures often degrade image quality during acquisition. To address these limitations, enrollment typically depends on specialized high-resolution scanners, which most existing enhancement methods are not designed to support. Consequently, identification rates for children remain significantly lower than those achieved with adult fingerprints. This study introduces Recursive Class Connectivity Classification (R3C), a novel framework that iteratively refines binary segmentation outputs from existing enhancement methods by extending ridge structures. R3C does not require modifications to the underlying classifier and operates without training data, which is not currently available for infant fingerprints. Instead, the method improves segmentation by repeatedly feeding the classified image back into the classification process, while combining each intermediate segmentation with the original input image. Experiments conducted on three fingerprint datasets using four different enhancement classifiers show that R3C can increase the True Acceptance Rate (TAR) by up to 4% for children and over 40% for newborns, compared to using the enhancement methods alone. A qualitative analysis further demonstrates that R3C reconnects fragmented ridge patterns, improving the visual quality of segmentation. Because it functions independently of the enhancement method used, R3C provides a flexible and broadly applicable solution for improving binary segmentation.

## I. INTRODUCTION

Biometric identification—or simply biometrics—is a widely adopted technology for identity validation that offers both convenience and security, as it relies on biological traits instead of password schemes. Fingerprints are a particularly popular biometric modality, and advancements in the field have made recognition possible with relatively inexpensive scanners [29]. Despite these innovations, significant progress in biometric systems for young children remains limited [38], hindering important humanitarian applications such as monitoring vaccination and nutrition campaigns and preventing baby swaps through maternity authentication [28], [38]. These issues are most pronounced in infants, who represent the most error-prone age group [20].

Differences in fingerprint size are the most critical issue in this subject. Infant fingerprints are significantly smaller than their adult counterparts, which makes it more challenging to capture their features. High-resolution scanners—more than four times the standard 500 ppi used for adults—are often adopted to capture infant fingerprints [23], [25]. However, while hardware improvements allow for the capture of additional details, other factors such as the malleability and moisture of the skin of children can still blur and distort significant portions of the image [23], meaning that even with state-of-the-art scanners, infant fingerprints still lack the quality in structure definition observed in older individuals.

Figure 1 compares adult and infant fingerprints enrolled at 500 and 5,000 ppi, respectively. Despite the infant fingerprint being acquired at ten times the resolution, the adult example still exhibits more clearly defined patterns, as seen in the Region of Interest (ROI) segmentation. In addition to the size discrepancy, the challenges of infant fingerprints are further compounded by the malleability of infant skin and the individual's cooperation, which introduces additional challenges during acquisition and increases the risk of noise [23], [25]. These physiological and acquisition-related factors naturally lead to more consistent recognition performance for adult fingerprints compared to those of infants.

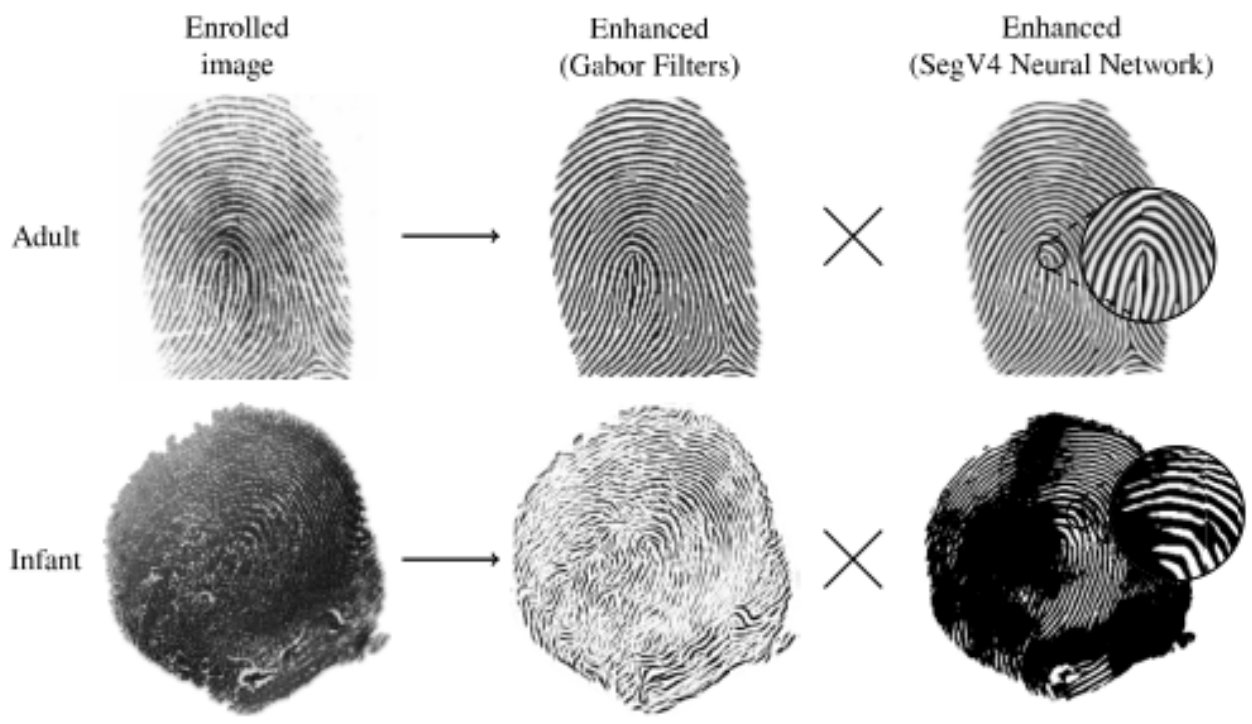

**FIGURE 1. Example of adult (500 ppi, FVC2002 database [4]) and infant (5,000 ppi, UTFPR-NFD database [37]) fingerprints enhanced with Gabor filters [3] and a commercial Neural Network [36].**

To improve the matching potential of enrolled fingerprints, image enhancement has become a standard practice in biometric systems [29], aiming to increase the contrast between ridge and valley patterns, which typically produces a binary segmentation. However, in infant fingerprints, the narrower spacing between ridges and valleys results in a blurrier appearance, limiting the effectiveness of conventional enhancement methods. As shown in Figure 1, Gabor filters successfully extract continuous ridge flow for the adult fingerprint, but capture few, if any, meaningful structures of the infant sample.

Despite the limitations of existing enhancement methods, few studies have focused specifically on ridge and valley segmentation in infant fingerprints. Filter-based approaches have produced mixed results, often introducing visual artifacts in fingerprints from young children [26]. Deep learning techniques have shown promise through super-resolution methods [27], [28], [33], [35], shifting focus from binary segmentation to image upscaling and novel matching pipelines. Commercial solutions have also emerged, such as SegV4, an experimental Convolutional Neural Network (CNN) proposed by [36]. However, as shown in Figure 1, the CNN also struggles to capture the full ridge flow in infant fingerprints, with low-quality regions often appearing as undifferentiated dark blobs.

To address the challenges of infant fingerprint enhancement, this study introduces the Recursive Class Connectivity Classification (R3C), a novel framework designed to improve the outputs of binary segmentation methods. The core idea is to recursively propagate previously segmented (*i.e.*, classified) pixels to guide subsequent predictions, following the concept of "pixel connectivity" in segmentation as originally defined by [24] in the ELEMENT framework. While ELEMENT was developed for retinal vessel segmentation, R3C generalizes and adapts this principle to fingerprint segmentation, offering simpler integration with a wide range of binary classification tasks.

Experimental results show that this novel approach can increase the matching rates of existing enhancement methods by up to 4% for older children and 40% for newborns, relative to standalone application of these methods, without requiring any modification to the underlying classifiers. Additionally, a complementary qualitative analysis reveals that R3C effectively extends enhancement coverage to fingerprint regions previously missed by the original classifiers. Nonetheless, absolute identification rates remain low, especially for newborns, and in some cases R3C is shown to introduce additional errors by propagating false ridge structures.

In the following, Section II discusses the related work. Section III details the proposed framework. Section IV compares its performance with other enhancement techniques over three infant fingerprint datasets. Section V discusses the implications of the findings. Lastly, Section VI summarizes the contributions and outlines future directions.

## II. LITERATURE REVIEW

This section presents key fingerprint enhancement works relevant to this study, followed by a discussion on pixel connectivity in segmentation, a central concept that underpins the design of the proposed framework.

### A. FINGERPRINT ENHANCEMENT

Traditional computer vision techniques have been extensively explored for adult fingerprint enhancement. One of the most prominent approaches is the use of Gabor filter banks, first introduced by [2] and [3], to enhance ridge patterns in standard 500 ppi fingerprints. This method proved effective in masking low-quality areas considered "extremely harmful for minutiae extraction" [2]. However, the reliance on orientation-sensitive filters often led to inaccuracies near critical singular points [6]. To address these challenges, a bandpass filter can be used in adult fingerprints [6].

Frequency-based probabilistic methods [7], combined with Gaussian window functions over overlapping blocks [15], have been used for ridge segmentation and the correction of discontinuities. On the topic of ridge discontinuities, [11] proposed an iterative method that applies Gabor filters to high-quality seed points and progressively extends enhancement to lower-quality regions. More recently, [21] introduced a fingerprint quality measure coupled with an iterative enhancement approach that uses a mixture of fingerprint features to estimate localized quality and guide Gabor filtering in restoring degraded blocks.

However, none of the aforementioned methods were originally designed for infant fingerprints, and the effectiveness of enhancement techniques in this context remains an open question in the biometric literature [38]. Instead, infant biometric studies have typically focused on the challenges of acquiring high-quality images [23], [25] and on the development of novel matching schemes [22], [28].

For enhancement, Deep Learning has emerged as the most common approach. Nonetheless, it is primarily used for super-resolution [28], [33], [35], rather than binary segmentation [27]. In the application of super-resolution to infant fingerprints, [28] improve identification by combining minutiae, texture, and latent matchers, incorporating a fine-tuned Residual Dense Network (RDN) and a minutiae scaling model that simulates aging. Koop et al. [33] employ a similar RDN to upscale high-resolution (3,000 ppi) infant fingerprints, which are then enhanced using a commercial CNN [36]. Machado et al. [35] train a visual transformer to perform super-resolution on 500 and 3,000 ppi infant fingerprints, using the Fast Fourier Transform (FFT) enhancement from NIST Biometric Image Software (NBIS) package [8] on the upscaled images, which must be downsampled to 500 ppi due to NBIS limitations. Shi and Liu [27], in contrast, propose a proprietary dense pyramid CNN with a minutiae attention block to jointly perform super-resolution and binary segmentation of 500 ppi infant fingerprints.

The focus on super-resolution reflects broader trends in image enhancement. However, the application to infant fingerprints remains limited by the scarcity of training data. Consequently, more robust filtering techniques become attractive alternatives, particularly as recent approaches have even been shown to surpass Neural Network models in some general-purpose resolution enhancement tasks [34].

The effectiveness of latent fingerprint enhancement methods in the infant domain remains an open question in the literature. Morphological differences, such as narrower ridge-valley spacing and lower overall contrast in infant fingerprints, raise concerns about the application of these approaches across domains. Deep learning is also prevalent in this context, with state-of-the-art methods ranging from the integration of Gabor filters into CNN architectures [19], to the use of Generative Adversarial Networks (GANs) for enhancement generation [30], and CNN autoencoders as filter prediction models that reconstruct degraded ridge patterns [32].

Table 1 summarizes key fingerprint enhancement studies relevant to this work. While traditional approaches have largely relied on filtering techniques [3], [7], recent efforts to improve latent and infant fingerprints are increasingly based on deep learning methods [19], [27], [30], [32]. In particular, Shi and Liu [27] remains the only prior work that introduces a dedicated model for enhancing infant fingerprints. However, such is limited to standard-resolution (500 ppi) images of children aged 3 months to 3 years and is evaluated solely on a private dataset.

The present study contributes both a novel method for refining enhancement outputs in infant fingerprints and a comparative evaluation of various enhancement techniques across multiple datasets, including two publicly available infant fingerprint databases [17], [22] and one private newborn dataset [37]. This work builds upon our previous proposal, which leveraged unconstrained iterations for data augmentation [31].

### B. PIXEL CONNECTIVITY FOR SEGMENTATION

Pixel connectivity is the basis of the method proposed by this work. The concept, with respect to binary classification, was initially introduced by [24] as a region-growing strategy for enhancing eye-vessel segmentation. The original framework (ELEMENT) combines features extracted by multiple classifiers to guide a region-growing process that expands vessel segmentation from a set of seed points. This iterative process continues as long as new seed points can be selected, progressively refining the classification of both vessel and background pixels.

For binary segmentation problems, connectivity is defined by [24] as the combination of two features: (i) immediate connectivity and (ii) radial connectivity. Both features are incorporated into a probability function $P$, defined in Eq. (1), which varies based on the number of already classified vessel pixels in proximity to the candidate pixel $I_{i,j}$, and differ only in the size of the neighborhood that is considered.

$$P(I_{i,j}) = \frac{e^{-M_{i,j}} - 1}{Z} \tag{1}$$

where $Z$ is a normalization constant, and $M_{i,j}$ represents the neighborhood around the candidate pixel. Essentially, connectivity is a measure of the correlation between a pixel and its neighbors. As such, $P$ estimates the probability that $I_{i,j}$ belongs to the vessel structure, supporting both the region-growing segmentation process and the suppression of incorrectly segmented background regions.

However, the reliance on seed point selection poses a significant challenge for the integration of ELEMENT with infant fingerprint enhancement. In the domain of fingerprints,

**TABLE 1. Related Fingerprint Enhancement Works.**

| Study | Enhancement | Databases | Age | Type |
|---|---|---|---|---|
| Hong, Wan, and Jain [3] | Gabor | MSU (no citation) | Adult | Ridge |
| Chikkerur, Cartwright, and Govindaraju [7] | STFT | FVC2002 DB3 [4] | Adult | Ridge |
| Tang et al. [19] | CNN | NIST SD27 [10] & FVC2004 [5] | Adult (latent) | Ridge |
| Shi and Liu [27] | CNN | Closed | Adult (training) & 3m. - 3y. | SR & Ridge |
| Zhu, Yin, and Hu [30] | GAN | NIST SD14 [10] (training) & NIST SD27 [10] & IIIT-Delhi MOLF [13] | Adult (latent) | Ridge |
| Kriangkhajorn, Horapong, and Areekul [32] | CNN | NIST SD27 [10] & IIIT-Delhi MOLF [13] IIITD-MFLSD [14] % NIST-SD302 [18] | Adult (latent) | Ridge |
| R3C (this work) | Gabor & FFT & CNN & GAN | Closed & CMBD [17] & NITG [22] | 3h. - 4y. | Ridge |

**STFT: Short-time Fourier Transform*
**FFT: Fast Fourier Transform*
**SR: Super-resolution*

ridges and valleys (Level 1 features) form the basis for extracting other important features (such as minutiae) [38]. Consequently, when the extraction of these basic structures is inconsistent, the automated selection of seed points becomes a major barrier. The following section introduces the proposed method, which addresses these limitations by leveraging the tendency of classifiers to iteratively connect and refine fingerprint enhancement through the use of their own previous predictions.

## III. PROPOSED METHOD

This section introduces the Recursive Class Connectivity Classification (R3C) method proposed in this study to enhance fingerprint image segmentation. Section III-A presents the overall framework of R3C, while Section III-B details the output regularization approach designed to minimize noise in the fingerprint enhancement improvements produced by R3C.

### A. R3C

R3C is a framework designed to enhance binary segmentation without requiring any modification to classifiers or input data. In the proposed method, the seed points and explicit connectivity features of ELEMENT [24] are omitted. Instead, region growing is achieved by repeatedly incorporating the classifier's output back into the original input image, thereby iteratively refining the binary segmentation as the output progressively improves. This allows class information to be propagated by reinserting and recombining the classified output with the original image.

By eliminating the explicit reliance on connectivity and seed points, R3C also enables seamless integration across a wide range of image domains with minimal effort. In other words, the proposed framework takes advantage of the concept of local neighborhood context as an indicator of pixel class, along with the continuity of segmented structures, such as eye vessels or ridges and valleys of fingerprints, by combining the complete segmentation output predicted by domain-specific classifiers with their original input.

Given an input image $M$, the respective output of a binary classification function is $C = f(M)$. The combination of these images ($M$ and $C$) is performed through a parameterized weighted sum, defined by

$$A = M + \alpha \times C \tag{2}$$

where $\alpha$ is a constant that modulates the classifier output $C$, effectively controlling its transparency when blending into the input $M$. This operation resembles an image overlay, with $\alpha \times C$ superimposed on $M$. In R3C, this operation is executed recursively: $M$ is repeatedly blended with $C$, and the resulting combination ($A$) serves as input for the next iteration.

An additional parameter $\gamma$ could be introduced as a decay factor to modulate $\alpha$, gradually reducing its value until it naturally terminates the recursive process at $\alpha = 0\%$. This approach could also favor earlier segmentation outputs, as the influence of newly generated predictions would diminish over time due to increasingly subtle blending. However, in this study, the recursion is instead controlled by a stopping condition defined by Eq. 3 in which $\epsilon$ is a constant denoting the predefined minimum threshold for the difference rate $d(C_i, C_{i-1})$.

$$d(C_i, C_{i-1}) \leq \epsilon \tag{3}$$

For the experiments in this study, the difference function is a binary segmentation index defined in Eq. 4 with $\sum C$ representing the sum of segmented pixels in $C$. The resulting value describes the percentage of segmentation added through recursion at iteration $i$ in relation to the previous step, $i - 1$.

$$d(C_i, C_{i-1}) = \frac{\sum C_i - \sum C_{i-1}}{\sum C_i} \tag{4}$$

Until the stopping condition is triggered, R3C continuously feeds $A$ as the input for the next iteration step $i + 1$. Once $d(C_i, C_{i-1})$ becomes equal to or falls below the threshold $\epsilon$, R3C concludes processing and returns $C_i$. This means that the final output of the proposed framework is a binary segmentation image that has been iteratively refined by the classifier function $f(M)$ through the manipulation of inputs and outputs defined in Eq. (2). Figure 2 presents a visual overview of how this pipeline affects inputs and outputs.

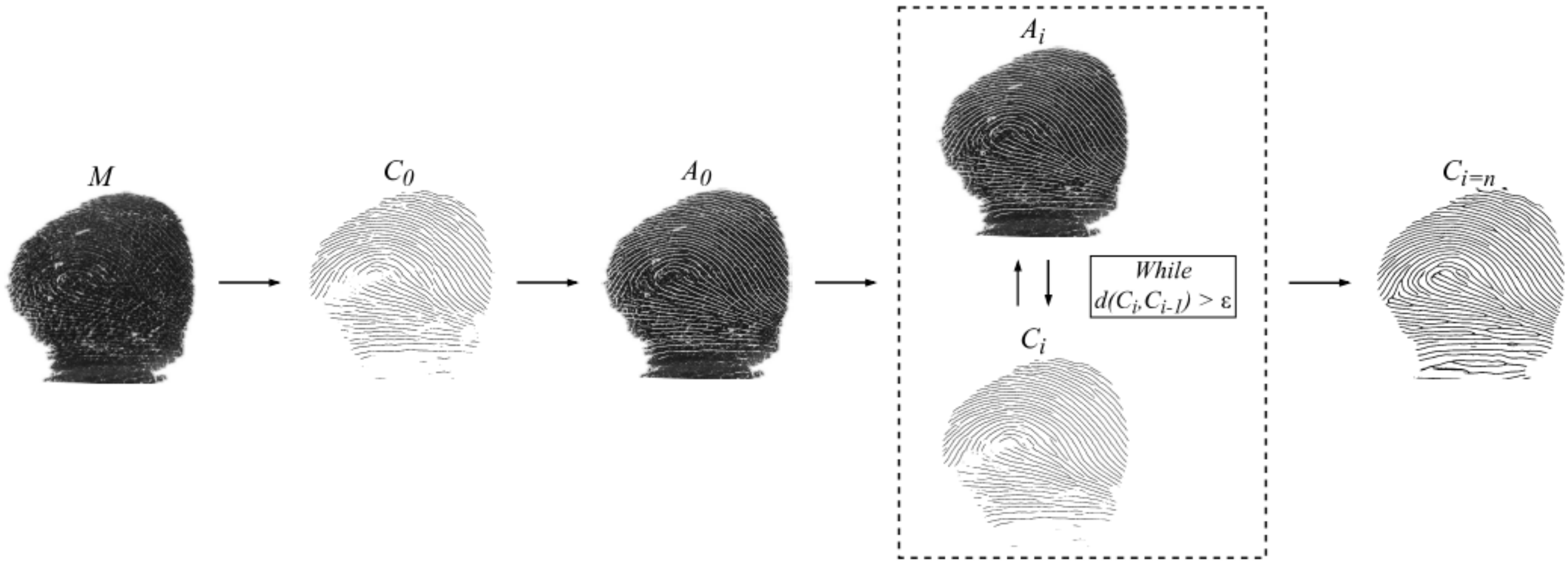

**FIGURE 2. Overview of the image transformations that occur during the R3C pipeline. The recursive process, highlighted by the dotted box, continues until an iteration $n$ where $d(C_i, C_{i-1}) \leq \epsilon$.**

It is important to note that R3C depends on two key considerations. First, as Figure 2 illustrates, an initial enhancement of the input image $M$ is always required before the recursive process can begin. Consequently, for the initial step ($i = 0$), the stopping criterion is not applied; instead, the method proceeds directly to the next iteration using $A$ as the input. As a result, in the best-case scenario R3C executes only two iteration steps.

Second, $A$ is progressively modified throughout the recursive iterations. As defined by Eq. (2), $A$ is a combination of the classifiers input and output images, which R3C re-utilizes as the input for the next epochs. As such, the segmentation output of the classifier is repeatedly stacked on top of the original image $M$ to ensure a stable convergence of $d(C_i, C_{i-1})$ towards the predefined threshold $\epsilon$, as the classifiers tend to maintain most of the structures from previous predictions (Section IV-E further details the behavior of the classifiers utilized in this study). The procedure followed by an R3C iteration is formalized in Eq. 5.

$$R3C(A, C_{i-1}) = \begin{cases} A + \alpha \times C_i, & \text{if } d(C_i, C_{i-1}) > \epsilon \\ C_i, & \text{otherwise} \end{cases} \tag{5}$$

The proposed method can be separated into two functions: (i) an initialization function that receives an image $M$ and computes the initial variables $A$ and $C$ (Algorithm 1) at $i = 0$, and (ii) the recursive improvement function, which continuously combines its input image $A$ and the output of the classifier function $C_i$ until the stopping criterion is met (Algorithm 2). Once $d(C_i, C_{i-1}) \leq \epsilon$, $C_i$ is returned as the final refined segmentation of $M$.

### B. OUTPUT REGULARIZATION

In order to regularize the output $C$ of R3C, the combination defined in Eq. 2 was modified to utilize a skeleton of the classifier prediction. This skeleton is obtained by applying the Zhang and Suen thinning algorithm [1] to the classifier output $f(C)$. This step ensures that the recursive improvement preserves the width of the ridges and valleys present in the original prediction. Without this regularization, the iterative expansion inherent to R3C can introduce noise by widening the segmented structures beyond their intended boundaries.

Furthermore, the integration of R3C with the filter-based fingerprint enhancement methods utilized in Section IV required an additional processing step before thinning could be applied. As these enhancement methods are designed to highlight fingerprint ridges, they generate outputs with white

**Algorithm 1** `improve_classification`: Initializes Binary Segmentation and Prepares for Recursive Improvement

**Require:** Input image $M$
**Ensure:** Final binary segmentation $C$
1: **function** `improve_classification`($M$)
2: $C_0 \leftarrow f(M)$
3: $A_0 \leftarrow M + \alpha \times C_0$
4: **return** `recursive_improvement`($A_0$, $C_0$)

**Algorithm 2** `recursive_improvement`: Recursively Applies Pixel Connectivity by Feeding the Updated Segmentation Back Into the Classifier

**Require:** Composite image $A$, previous segmentation $C_{i-1}$
**Ensure:** Final binary segmentation $C_i$
1: **function** `recursive_improvement`($A$, $C_{i-1}$)
2: $C_i \leftarrow f(A)$
3: **if** $d(C_i, C_{i-1}) \leq \epsilon$ **then**
4: **return** $C$
5: **else**
6: $A_{\text{new}} \leftarrow A + \alpha \times C_i$
7: **return** `recursive_improvement`($A_{\text{new}}$, $C_i$)
8: **end if**

ridges on a dark background. However, this is the inverse of the raw fingerprints ($M$), where ridges typically appear as dark regions due to their direct contact with the scanner surface [29], while valleys are captured as bright regions (essentially as background information). As Figure 3b illustrates, to reconcile this discrepancy and ensure compatibility with the assumptions of R3C, the output class labels were inverted so that bright valleys and dark ridges are represented as expected by the image combination process.

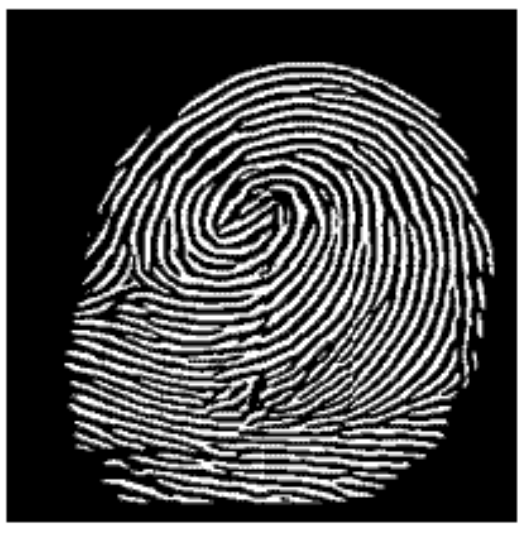

(a) Original Output (Ridges)

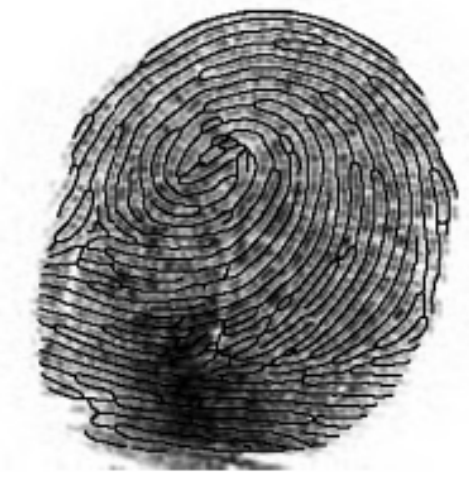

(b) Input + Valleys Composition

**FIGURE 3. Original ridge enhancement output (Figure 3a) and resulting Input and Valley composite image (Figure 3b). For illustration, the valleys were colored black (originally white).**

## IV. EXPERIMENTAL RESULTS

This section presents a comprehensive evaluation of the performance of various fingerprint enhancement methods for infants, and investigates how integrating the proposed R3C framework influences each method's effectiveness. Section IV-A introduces the databases used in this study, followed by an outlining of the enhancement methods and R3C integration details in Section IV-B. Section IV-C reports the fingerprint matching performance in different age groups of infants, and Section IV-D examines the matching performance of newborns during the first hours of life. Finally, Section IV-E provides a qualitative comparison of enhancement outputs, highlighting the improvements brought by the proposed R3C framework.

NIST's MindTCT and Bozorth3 [9] were used for minutiae extraction and matching, respectively. These are non-commercial tools that are approached as baseline solutions. The experiments aim to evaluate how different fingerprint enhancement methods (both standalone and integrated with R3C) perform in the context of infant biometrics. A 1:N matching protocol was adopted, where each probe was individually compared against the entire gallery of the corresponding dataset. No custom configurations or non-default options were applied to MindTCT or Bozorth3.

### A. INFANT FINGERPRINT DATASETS

- **CMBD [17]**: is an open multi-modal dataset for infant biometrics. It contains fingerprints of all ten fingers acquired from children aged 18 months to 4 years using a 500 ppi optical scanner. For this study, only the first session data was employed, comprising 1,190 genuine pairs. Images labeled "_1" were used as the gallery set, while those labeled "_2" served as probes for matching (2,380 images in total).
- **NITG [22]:** is an open dataset designed for infant fingerprint biometrics. It contains fingerprints from the thumbs of children (age undisclosed), acquired using a 500 ppi optical scanner, which comprises 1,102 genuine pairs. For this study, images labeled "_1" were used as the gallery set, while those labeled "_5" served as probes for matching (2,204 images in total).
- **UTFPR Newborns Fingerprint Dataset (UTFPR-NFD) [37]**: is a newborn fingerprint dataset assembled in a hospital environment using video-based acquisitions with 5,000 ppi optical scanners. The subjects are infants aged between 3 hours and 3 days after birth. In our study, only images from the thumbs and index fingers (of both hands) were utilized, as the capture protocol requires two acquisitions of these fingers to be performed during the same session. The fingerprint images employed for matching were extracted from video frames using a commercial fingerprint quality estimation Neural Network provided by the scanner manufacturer [36]. The final subset includes 377 genuine pairs (754 images), covering all hospital acquisitions from March to September 2024, except for some discarded as "in-error" (lacking paired thumb and/or index finger images). Images labeled "_1" were used as the gallery set, while those labeled "_2" served as probes for matching. The acquisition protocol has been approved with a Certificate of Presentation of Ethical Appreciation 73791023.7.0000.0177 at the Brazil Platform. Additional information about the dataset can be found in the pre-print document [37].

### B. FINGERPRINT ENHANCEMENT CLASSIFIERS AND R3C PARAMETERS

As a framework, R3C can be applied to any binary image segmentation domain simply by changing the classifier function $f()$. For our experiments, we employed four distinct fingerprint enhancement classifiers: (i) Gabor filters [3]; (ii) NBIS two-dimensional FFT implementation [8]; (iii) SegV4, a commercial U-Net CNN model specifically trained to enhance infant fingerprints [36]; and (iv) FingerGAN, a state-of-the-art adversarial network for latent fingerprint enhancement [30]. Table 2 presents the classifiers and their input limitations.

**TABLE 2. Fingerprint enhancement classifiers.**

| Classifier | Type | Shape | Intended for |
|---|---|---|---|
| Gabor [3] | Filter | None | 500 ppi adult |
| NBIS [8] | Filter | $500 \times 500$ | 500 ppi adult |
| SegV4 [36] | DLNN | None | 5,000 ppi infant |
| FingerGAN [30] | DLNN | $\geq 192 \times 192$ | 500 ppi adult/latent |

**DLNN: Deep Learning Neural Network*

The Gabor filter implementation utilized in this study does not impose input shape requirements, but its performance is optimized for images with a resolution of 350 × 350. Consequently, all images processed by this classifier were resized to 350 × 350 prior to enhancement. For the NBIS package, the images were subjected to a similar reshape before enhancement since its FFT implementation which does have a strict shape requirement of 500 × 500.

In contrast to the filter-based methods, the Neural Network classifiers exhibit challenges related to the scope of their training data rather than explicit input size constraints (aside from FingerGAN's minimum resolution of 192 × 192, which did not affect the images employed in this study). SegV4 was trained to enhance 5,000 ppi infant fingerprints acquired during the first months of life, which limits its applicability to fingerprints from older children, or those captured at lower resolutions. Conversely, FingerGAN was trained on 500 ppi latent (adult) fingerprint images, making it well-suited for degraded fingerprint data but not specifically adapted to handle infant or high resolution fingerprints.

As a flexible framework, R3C did not required any modifications to either the underlying classifiers or its recursive pipeline. However, the framework still necessitates some adjustment of its internal parameters, $\alpha$ and $\epsilon$.

The blending parameter $\alpha$ was optimized independently for each classifier and dataset. This optimization involved evaluating the performance on 100 genuine pairs from the CMBD database, 100 genuine pairs from the NITG database, and 50 genuine pairs from the UTFPR-NFD database (as it contains fewer samples). R3C integration was tested using five distinct $\alpha$ configurations: very low (5%), low (25%), medium (50%), high (75%), and very high (100%). The optimal $\alpha$ value for each classifier and dataset was determined by selecting the configuration that yielded the highest True Acceptance Rate (TAR) at a False Acceptance Rate (FAR) of 0.1%, with FAR = 1.0% selected as a tiebreaker. TAR and FAR are related indexes used in biometric evaluation: TAR quantifies the proportion of correctly identified genuine matches, while FAR represents the percentage of impostor matches incorrectly accepted as genuine.

In contrast, the stopping condition threshold $\epsilon$ was fixed at 1% (i.e., a 0.01 difference). This value was chosen to enable meaningful segmentation growth while avoiding diminishing returns. For other applications, however, this threshold may be adjusted: higher values constrain the influence of R3C, resulting in more conservative region-growing, while lower values allow for more extensive segmentation expansion. To further ensure that the segmentation had truly stabilized, the method was also modified to terminate the recursive process only if the stopping condition was satisfied in two consecutive iterations.

As summarized in Table 3, the selected $\alpha$ values vary across classifier and dataset combinations. These variations are expected, given that each dataset comprises fingerprint images captured at different resolutions, from subjects of varying ages, and involving different sets of fingers—CMBD includes all ten fingers, NITG includes only thumbs, and UTFPR-NFD includes thumbs and index fingers. These anatomical and acquisition differences lead to variability in finger size and quality, which in turn may influence the matching performance and optimal configuration of R3C.

**TABLE 3. Optimized R3C $\alpha$ parameters.**

| Classifier | CMBD | NITG | UTFPR-NFD |
|---|---|---|---|
| Gabor + **R3C** | 100% | 5% | 25% |
| NBIS + **R3C** | 75% | 100% | 25% |
| SegV4 + **R3C** | 75% | - | 50% |
| FingerGAN + **R3C** | 100% | 100% | 75% |

**TABLE 4. TAR for different FAR values (CMBD).**

| Enhancement | FAR=0.1% | FAR=1% | FAR=10% |
|---|---|---|---|
| Gabor | 32.521 | 43.95 | 63.277 |
| Gabor + **R3C** | 33.193 | 45.21 | 64.706 |
| NBIS | 31.008 | 41.261 | 60.504 |
| NBIS + **R3C** | 32.773 | 43.025 | 62.101 |
| SegV4 | 1.765 | 5.798 | 19.328 |
| SegV4 + **R3C** | 2.773 | 6.387 | 20.084 |
| FingerGAN | 35.63 | 47.059 | 66.134 |
| FingerGAN + **R3C** | 28.067 | 39.748 | 61.092 |

**CMBD: 2,380 images (1,190 pairs)*

Notably, due to resolution and possibly subject age, the SegV4 classifier was unable to enhance the NITG dataset images, producing blank outputs regardless of R3C integration. Consequently, this classifier was not employed for the NITG dataset.

Moreover, the integration of FingerGAN with R3C required a 3× down-sampling of the UTFPR-NFD inputs to reduce computational load. As implemented, R3C necessitates a minimum of three enhancement calls per image, which extended FingerGAN's execution time to several minutes per image (at 5,000 ppi) when using a batch size of 256 on an NVIDIA A100 GPU (40 GB). Importantly, this down-sampling was only applied during the R3C-integrated execution and did not affect the standalone FingerGAN enhancement for this dataset.

An additional class/color inversion was also required in order for MindTCT to correctly interpret the outputs produced by FingerGAN. This inversion is essentially the reverse of the one applied to the filter-based methods for R3C integration. This adjustment significantly improved FingerGAN performance when paired with Bozorth3 for matching.

### C. FINGERPRINT ENHANCEMENT FOR CHILDREN

Table 4 presents the TAR achieved by the evaluated enhancement methods on the CMBD dataset. Results are shown for both the standalone enhancement methods and their corresponding versions integrated with the proposed R3C framework (denoted as *+ R3C*), across varying FAR levels.

Notably, the proposed R3C method improved the performance of Gabor filters, NBIS FFT, and the SegV4 CNN, yielding consistent increases in TAR across all evaluated FAR thresholds. The most significant relative improvement

**TABLE 5. TAR for different FAR values (NITG).**

| Enhancement | FAR=0.1% | FAR=1% | FAR=10% |
|---|---|---|---|
| Gabor | 58.984 | 68.966 | 81.034 |
| Gabor + **R3C** | 56.261 | 67.695 | 81.397 |
| NBIS | 53.721 | 67.967 | 83.757 |
| NBIS + **R3C** | 52.904 | 64.61 | 81.125 |
| FingerGAN | 45.735 | 54.174 | 67.332 |
| FingerGAN + **R3C** | 34.029 | 44.555 | 58.802 |

**NITG: 2,204 images (1,102 pairs)*

was observed in the SegV4 classifier at FAR = 0.1%, where TAR increased by approximately 57%. However, due to the resolution mismatch discussed in Section IV-B, the absolute gain remains modest—rising from 1.76% to 2.77%. For Gabor and NBIS, the most pronounced improvements occurred at FAR = 1.0%, with R3C integration resulting in relative gains of up to 4% over their standalone counterparts.

Despite these gains, FingerGAN remained the top-performing method overall, achieving TARs of 35.63% and 47.06% at FAR = 0.1% and 1.0%, respectively. These values surpassed the second-best performer (Gabor + R3C) by approximately 2.4 percentage points at both thresholds (TAR = 33.19% and 45.21%).

In contrast to the other enhancement classifiers, the integration of R3C with FingerGAN led to a noticeable degradation in TAR across all FAR thresholds. This decline is likely attributable to the introduction of segmentation noise, as later illustrated in Section IV-E, which may have increased the occurrence of false positives.

As shown in Table 5, overall performance on the NITG dataset is notably higher than on CMBD. The best-performing method, Gabor, achieved a TAR of 58.98% at FAR = 0.1%, which is nearly double the top score observed on the CMBD dataset (35.63%). This discrepancy may be attributed to differences in acquisition protocols, as NITG exclusively contains thumb fingerprint samples. Furthermore, higher scores may reflect a higher average age of the subject in the NITG dataset, although this cannot be confirmed, as specific age information is not disclosed.

Regarding R3C integration, TAR/FAR performance generally declined with the exception of a marginal gain observed for Gabor at FAR = 10% (81.40% with R3C, compared to 81.03% without). This further suggests that R3C can introduce noise that hampers performance at more restrictive FAR levels.

The Cumulative Matching Characteristic (CMC) curves presented in Figure 4 for CMBD (Figure 4a) and NITG (Figure 4b) corroborate the findings of the TAR/FAR analysis. In CMBD, FingerGAN consistently outperforms all other methods across all ranks. In contrast, results for the NITG dataset show a slight advantage for NBIS enhancement, which performs marginally better than both the R3C-integrated NBIS and the standalone Gabor filter approach.

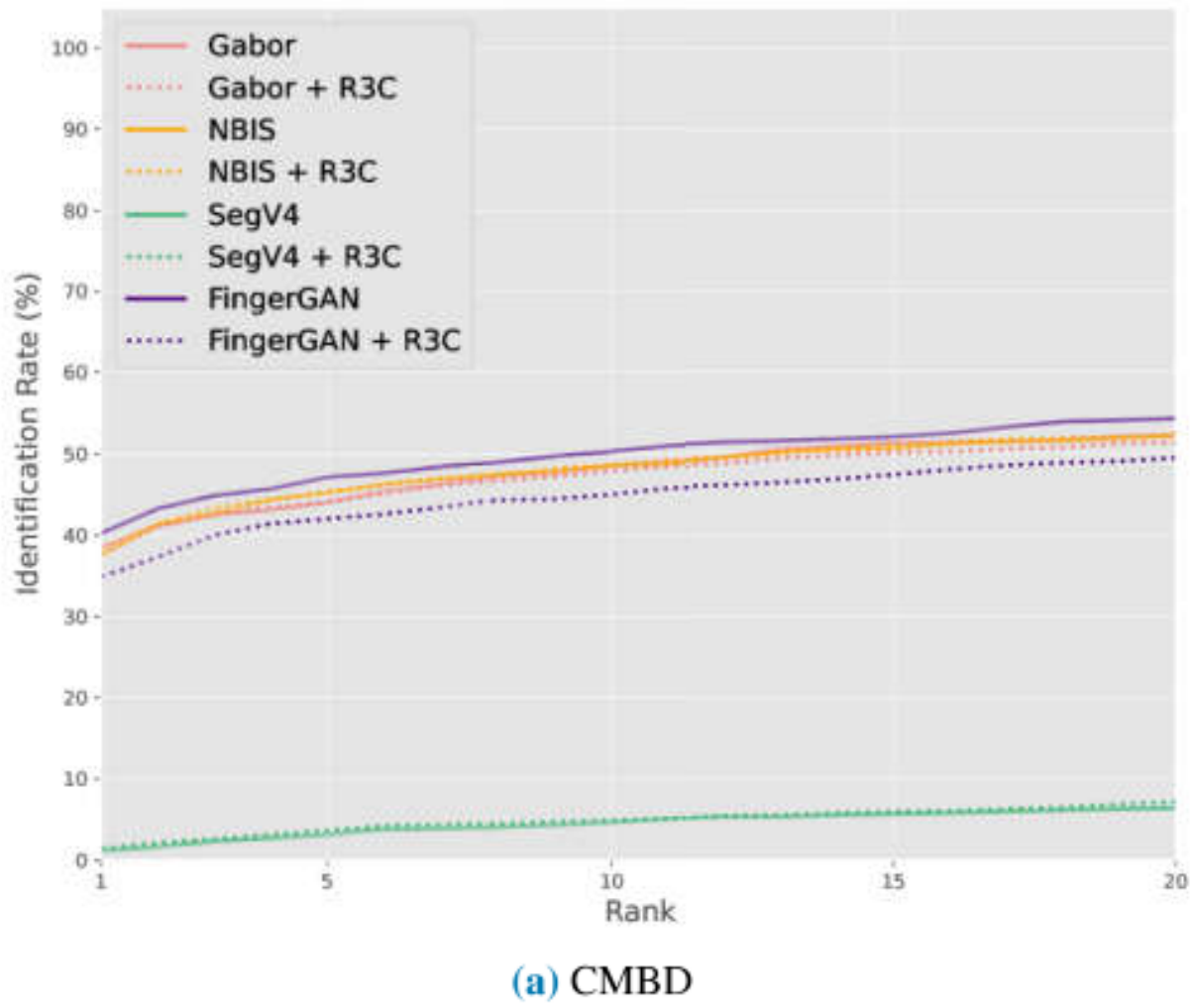

(a) CMBD

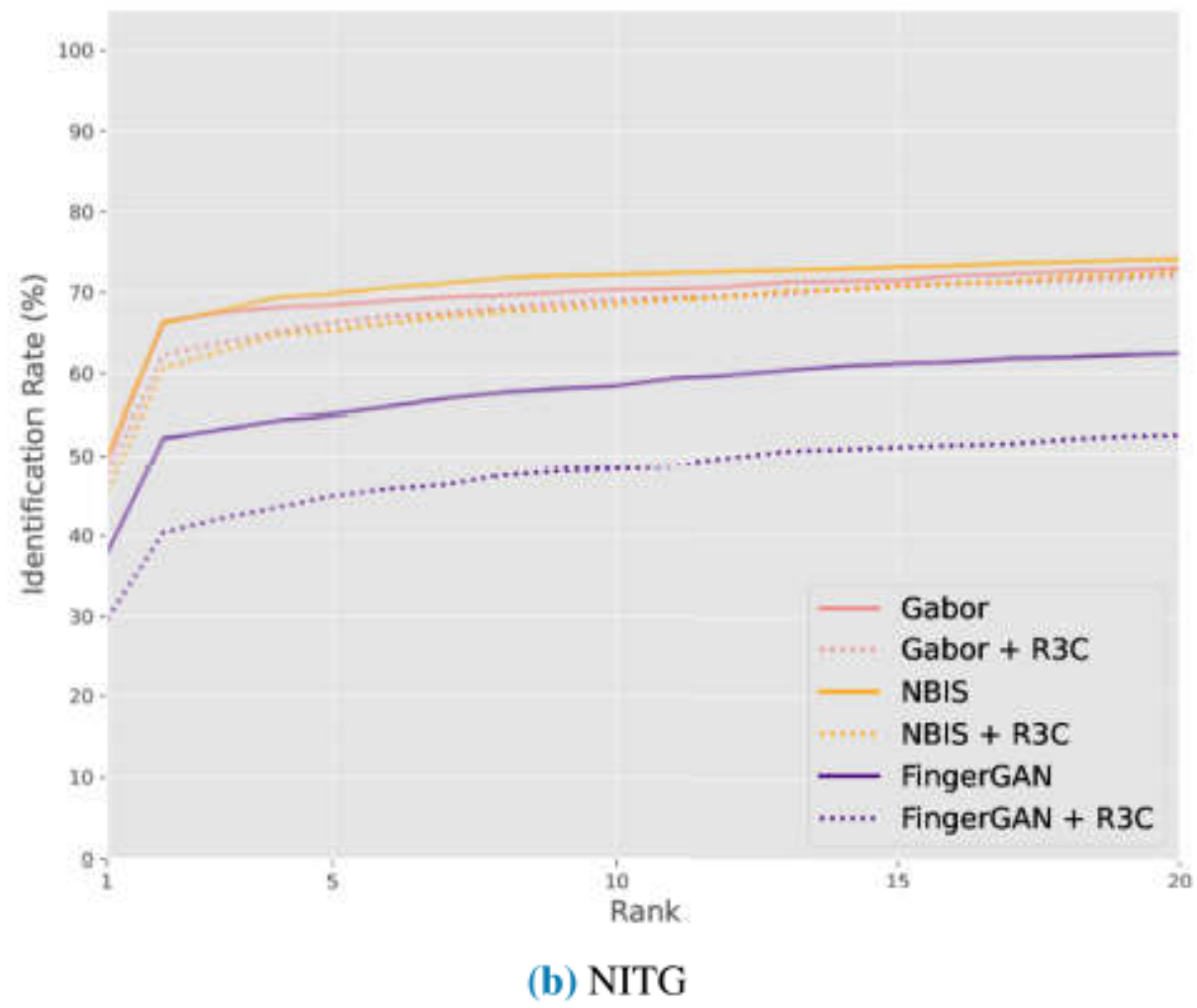

(b) NITG

**FIGURE 4. CMC curves for the CMBD and NITG datasets. Gabor is shown in red, NBIS in yellow, SegV4 in green and FingerGAN in purple. R3C integrated methods are shown as dotted lines.**

### D. FINGERPRINTS ENHANCEMENT FOR NEWBORNS

As shows Table 6, recognition performance is drastically reduced for newborn fingerprints. At FAR = 0.1%, the highest TAR was achieved by SegV4 and SegV4 + R3C, both reaching 1.592%. This stands in sharp contrast to results observed for older children, in which the lowest recorded TAR (apart from SegV4) was 31% for NBIS in CMBD.

Notably, SegV4 + R3C achieved the highest performance among all methods at low FAR levels, improving the standalone SegV4 TAR from 5.04% to 7.42% at FAR = 1.0% (a relative gain of approximately 47%). However, similar to the behavior observed with NBIS, the integration with R3C appears to have increased the occurrence of false positives, leading to a slight reduction in TAR at FAR = 10%.

The latent fingerprint enhancer, FingerGAN, failed to retrieve any correct matches at FAR = 0.1% when used in standalone mode, likely due to incomplete segmentation

**TABLE 6. TAR for different FAR values (UTFPR-NFD).**

| Enhancement | FAR=0.1% | FAR=1% | FAR=10% |
|---|---|---|---|
| Gabor | 0.265 | 2.122 | 10.345 |
| Gabor + **R3C** | 0.796 | 2.387 | 12.467 |
| NBIS | 0.265 | 1.592 | 14.854 |
| NBIS + **R3C** | 0.265 | 1.592 | 13.528 |
| SegV4 | 1.592 | 5.040 | 21.751 |
| SegV4 + **R3C** | 1.592 | 7.427 | 19.363 |
| FingerGAN | 0.000 | 1.070 | 14.439 |
| FingerGAN + **R3C** | 0.267 | 3.476 | 20.856 |

**UTFPR-NFD: 754 images (377 pairs)*

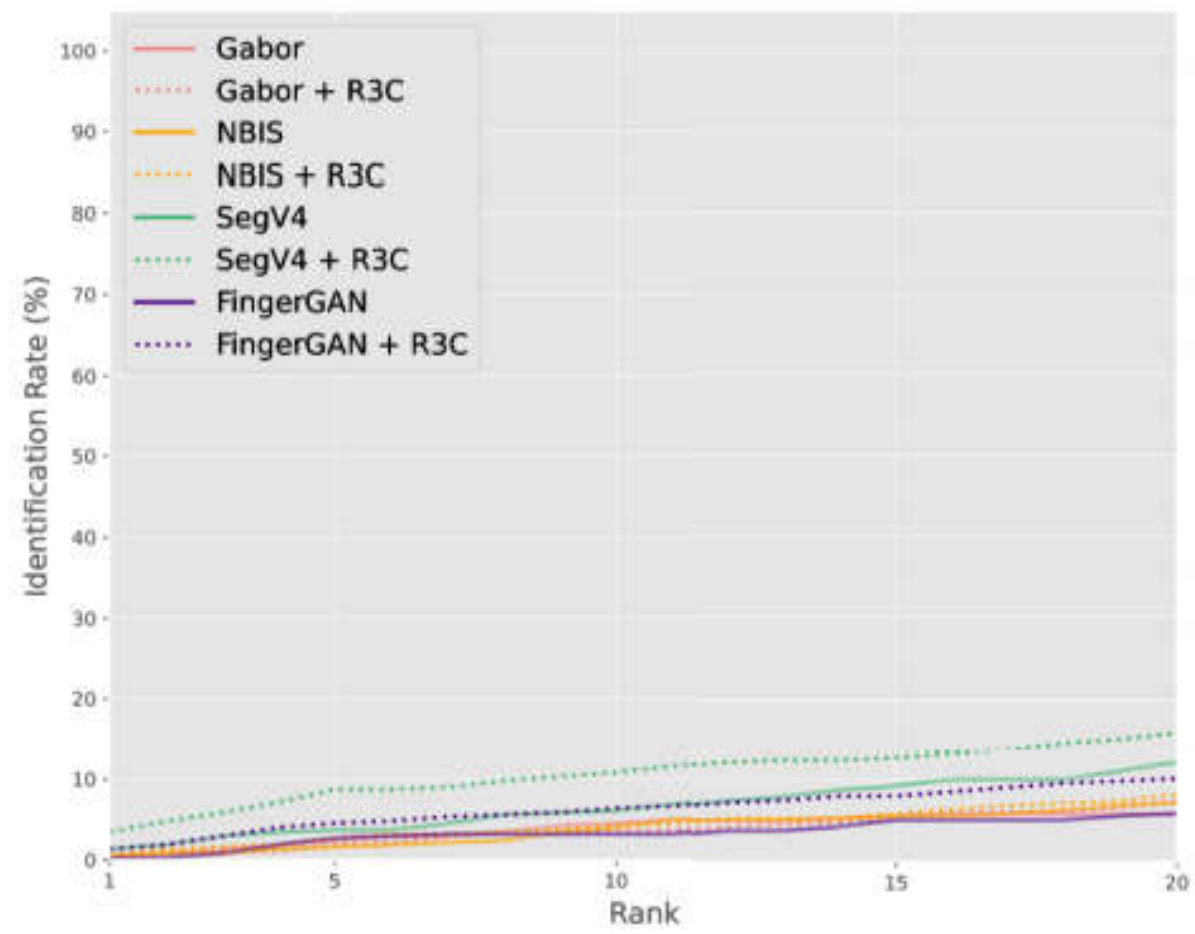

**FIGURE 5. CMC curves for UTFPR-NFD. Gabor is shown in red, NBIS in yellow, SegV4 in green and FingerGAN in purple. R3C integrated methods are shown as dotted lines.**

of ridges, as presented in Section IV-E. When integrated with R3C, FingerGAN exhibited a modest improvement, achieving a TAR of 0.27% at FAR = 0.1%. This improvement remained consistent across higher FAR thresholds, with TARs increasing to 3.48% at FAR = 1% and 20.86% at FAR = 10%, suggesting that R3C facilitated the recovery of additional ridge detail.

Similarly, the proposed method consistently improved Gabor filter performance across all thresholds. At FAR = 0.1%, R3C integration nearly tripled the TAR, increasing it from 0.27% to 0.79%, though the absolute gain was slight (approximately 0.53 percentage points). Nevertheless, the filter-based methods were generally outperformed by Neural Network approaches, with SegV4 (both standalone and R3C-integrated) and FingerGAN + R3C emerging as the more effective enhancement strategies for newborn fingerprints.

These findings are further corroborated by the UTFPR-NFD CMC curves presented in Figure 5, which show R3C consistently improving the rank-based identification performance of SegV4 while also enabling FingerGAN + R3C to reach comparable ranking levels. Nevertheless, the overall recognition performance for newborn fingerprints remained significantly lower than that observed for older children in the CMBD and NITG datasets, with even UTFPR-NFD Rank-20 values lower than CMBD Rank-1 results.

## *E. QUALITATIVE ANALYSIS*

The examples presented in Figures 6 and 7 illustrate how the evaluated enhancement methods perform across different age groups and demonstrate the impact of the proposed R3C integration on segmentation quality. For clarity, each figure contrasts the original fingerprint with the outputs of each standalone enhancement method, followed by their respective R3C-integrated results at three representative $\alpha$ configurations (5%, 50%, and 100%).

In the case of the CMBD sample (Figure 6), both filter-based methods produced visually complete initial segmentations, leaving limited room for R3C to make substantial modifications. Nonetheless, R3C still contributed to small but meaningful improvements. For example, in the highlighted region of the Gabor outputs, R3C successfully connected isolated pixel clusters near larger ridge structures—pixels that were detected by the filters but not sufficiently connected to form continuous features. These subtle refinements translated into measurable performance gains, with Gabor + R3C achieving a 2.86% improvement in TAR at FAR = 1.0%, as reported in Section IV-C.

The impact of R3C is particularly evident for the Neural Networks, as their initial output is less complete than the filter-based methods. SegV4, in particular, failed to produce adequate segmentation for this dataset, resulting in largely blank outputs with or without R3C integration. Conversely, FingerGAN showed substantial visual benefit from R3C. While the standalone FingerGAN output exhibited incomplete and fragmented ridge structures, higher $\alpha$ values (50% and 100%) expanded the segmentation to capture the entire central region and several peripheral areas. This progression clearly highlights how R3C can deliver both minimal refinements (at low $\alpha$ values) and extensive segmentation changes (at higher settings). However, despite these visual improvements, FingerGAN + R3C integration for CMBD did not translate into increased matching performance.

For the newborn fingerprints of UTFPR-NFD, Figure 7 illustrates how the completion behavior observed with R3C also extends to filter-based methods. For instance, the Gabor filters initially failed to enhance several fingerprint regions, producing outputs with large blank areas and isolated ridge fragments. Even at the lowest $\alpha$ setting (5%), R3C significantly improved ridge connectivity, progressively filling in these gaps and resulting in a more continuous and analyzable fingerprint structure.

Concerning NBIS, the visual impact of R3C is less pronounced. Since the FFT-based method already covered the entire fingerprint area, R3C's primary effect was to widen existing ridge structures, especially near the top and bottom regions of the image. This widening reduced scattered segmentation noise by reinforcing ridge continuity. However, these visual refinements did not translate into measurable performance gains; NBIS + R3C showed no change in TAR at FAR levels of 0.1% and 1.0%.

As with the CMBD sample, the impact of the proposed method was more pronounced for the Neural Networks.

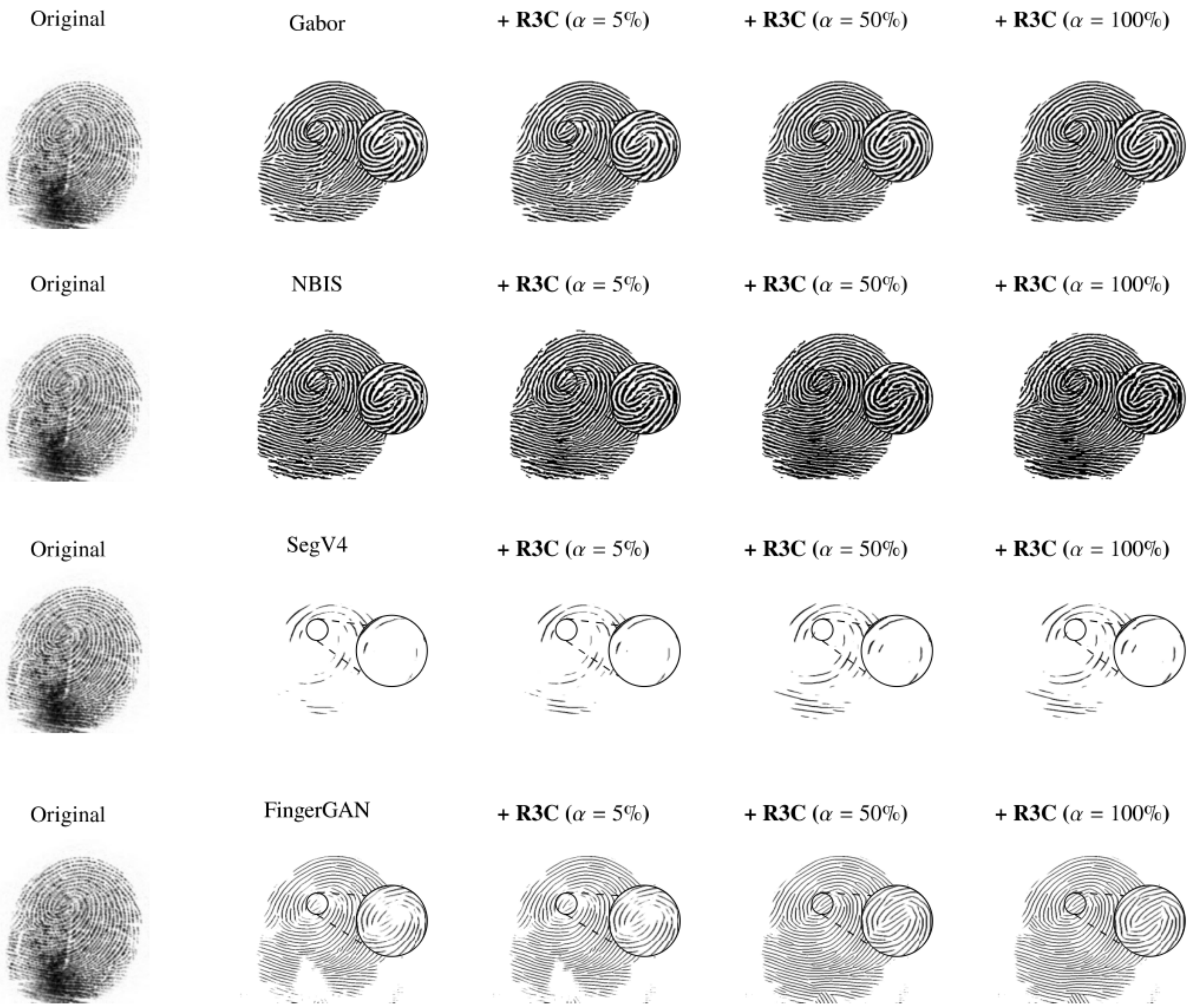

**FIGURE 6.** Comparison of enhancement methods on a CMBD sample. For demonstrative purposes, the output colors of Gabor, SegV4 and FingerGAN were inverted, and the structures segmented by FingerGAN were dilated.

SegV4 outputs are notably improved, with R3C connecting fragmented ridge structures while preserving the original ridge flow, leading to substantial gains in TAR at both low FAR thresholds. FingerGAN also benefited from R3C in visual appearance and matching performance. However, the observed improvements must be viewed in the context of the very poor standalone FingerGAN output, which was highly incomplete for this dataset. At lower $\alpha$ levels, the R3C-enhanced FingerGAN output resembled Gabor-like ridge patterns. In contrast, higher $\alpha$ values (*e.g.*, 100%) introduced dense, noisy connections with frequent abrupt orientation changes—likely a consequence of the 3× down-sampling applied during R3C processing to reduce computational load.

Notably, R3C's sensitivity to $\alpha$ depends on the enhancement classifier's behavior, which the qualitative analysis shows to vary across datasets. For instance, the Gabor filters' output for the CMBD sample presented only minor alterations even at the maximum $\alpha$ value (Figure 6); consequently, the optimized parameter for this method/dataset pair was 100% (Table 3). In contrast, for UTFPR-NFD, substantial changes occurred even at the minimum $\alpha$ value (Figure 7), with the parameter search yielding an optimal value of 25%. Thus, future applications of the proposed framework are advised to implement per-dataset tuning to avoid sub-optimal configurations that can either introduce errors (high $\alpha$) or limit improvements (low $\alpha$).

Regardless of limitations, R3C integration allowed FingerGAN to emerge as one of the best performers for UTFPR-NFD, emphasizing the potential of latent enhancement methods for infant fingerprint biometrics.

## V. DISCUSSION

The enhancement of infant fingerprints appears to be largely dictated by the age of the enrolled subjects. Filter-based methods such as Gabor and FFT produce relatively complete enhancements for 500 ppi images of older children (typically 18 months of age or older). However, for newborns, even the added detail provided by 5,000 ppi acquisitions is insufficient to close the performance gap without the use of specialized enhancement techniques. This ''age effect'' is well documented in the literature [12], [16], [20], [38], and efforts such as the U-net model [36] used in this study can significantly increase rates for younger subjects.

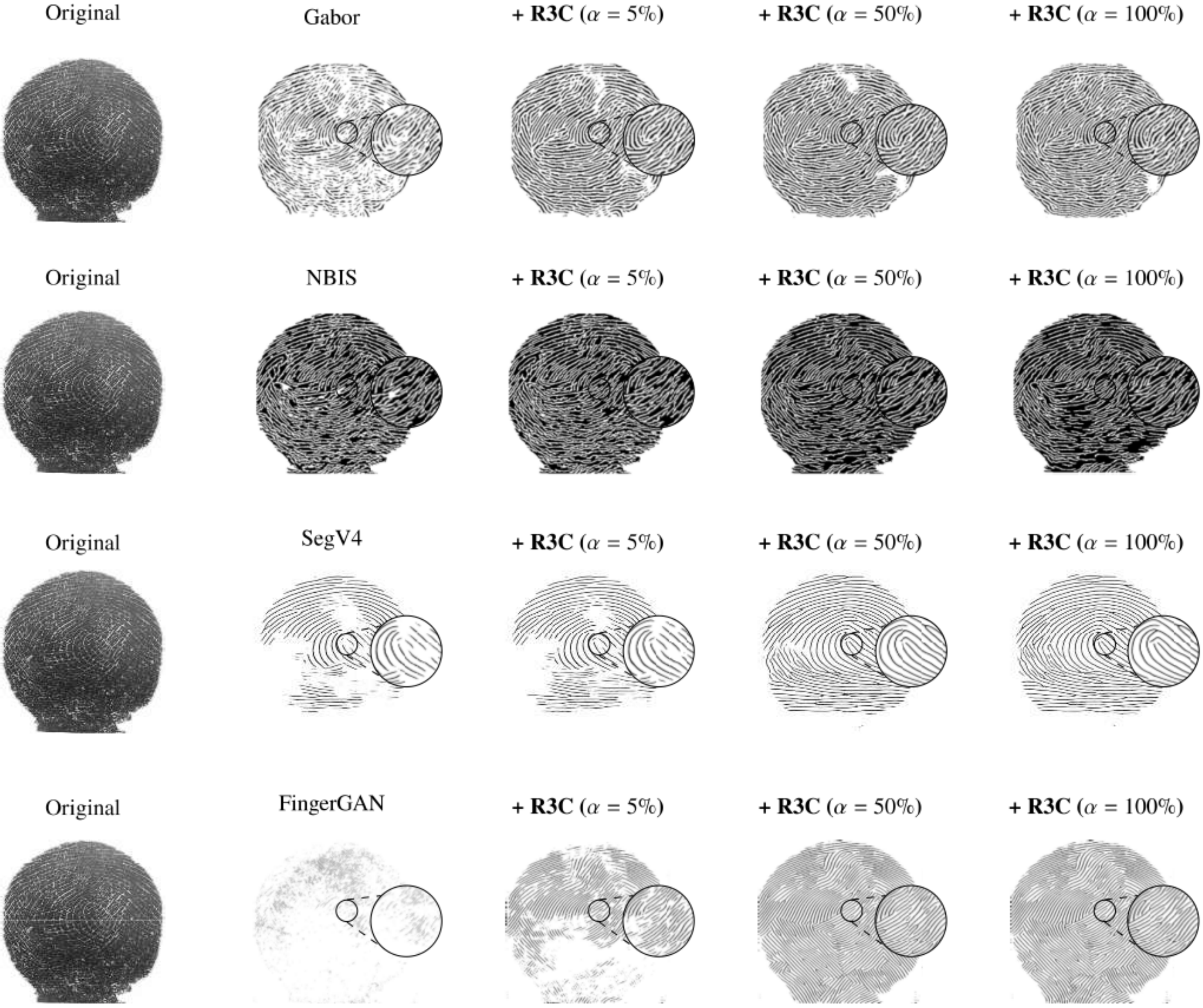

**FIGURE 7. Comparison of enhancement methods on a UTFPR-NFD sample. For demonstrative purposes, the output colors of Gabor, SegV4 and FingerGAN were inverted, and the structures segmented by FingerGAN were dilated.**

Latent enhancement methods are similarly promising, particularly intelligent models such as FingerGAN [30], which offer the potential for fine-tuning and targeted performance gains. However, progress in this area remains hindered by limited data availability, especially the lack of annotated datasets with expert-validated ground truth outputs, which prevents broader and more effective application of these specialized approaches.

The results in Section IV show that R3C can enhance segmentation output without requiring training data. For the CMBD dataset, R3C provided modest yet consistent improvements for filter-based methods, while delivering more significant gains for deep learning models on the UTFPR-NFD dataset. Still, these improvements do come at a significant computational cost, particularly when applied in conjunction with more complex models like GANs. Additionally, while R3C generally improves segmentation, it can also introduce false ridge structures that increase error as observed with FingerGAN on the CMBD and NITG datasets. Visual inspection (Figure 6) reveals that while the ridge structure at $\alpha = 100\%$ resembles Gabor and NBIS outputs more closely than the original enhancement, errors arise in minutiae points. For instance, a spurious bifurcation appears in the upper center of the $\alpha = 100\%$ output that is absent at $\alpha = 75\%$. This artifacts likely stem from FingerGAN's tendency to produce progressively sharper ridge angles as $\alpha$ increases, evident in the center/base of the CMBD sample (Figure 6) and throughout the UTFPR-NFD output (Figure 7).

Perhaps most importantly, the qualitative analysis in Section IV-E highlights how R3C can improve segmentation coverage in a way that also benefits the annotation process. By converting what would otherwise be a complex and time-consuming manual segmentation task into a simpler correction process focused mainly on erasing incorrect structures rather than adding missing ridges and valleys, R3C holds promise for both research and operational workflows.

Regarding matching performance, prior studies have shown that finger fusion [17], [28], which involves combining multiple acquisitions from different fingers, is particularly effective for children. Similarly, the use of latent matchers [28] has been shown to boost recognition rates, and commercial off-the-shelf (COTS) matchers may also improve absolute matching performance for children. Nevertheless, the fundamental goal of enhancement remains the same: to improve the matching potential of individual samples. Consequently, innovations in classifiers and matching algorithms will continue to be essential for achieving reliable identification performance in infant biometrics.

## VI. CONCLUSION

Infant biometrics presents recognition rates significantly lower than those achieved for adults. The primary challenges arise during enrollment, where the small size, softness, and malleability of children's fingerprints result in inconsistent ridge patterns, increased noise levels, and limited data quality. Compounding these issues is the lack of publicly available, high-quality datasets for developing and evaluating fingerprint recognition systems targeting young subjects.

To help address these limitations, this study introduces R3C, a Recursive Classification Connectivity framework designed for the iterative refinement of binary fingerprint segmentations. R3C enhances segmentation consistency and connectivity through repeated classification iterations without relying on predefined connectivity features or manually selected seed points. Its design enables the extension of classifier predictions to reduce enhancement noise and improve the coverage of low-quality regions often left blank by baseline methods.

The proposed method was evaluated across two public child fingerprint datasets [17], [22] and one proprietary newborn database [37], using four distinct enhancement classifiers [3], [8], [30], [36] spanning adult, infant, and latent fingerprint domains. Results demonstrate that R3C can improve the performance of these enhancement approaches, with relative TAR increases of up to 4% for older children and over 40% for newborns. However, absolute identification rates remain modest, as top performers at FAR $=$ 1.0% reached 68.97% TAR for the NITG dataset (undisclosed age range), 47.06% for CMBD (18 months to 4 years), and just 7.43% for newborns in UTFPR-NFD. Furthermore, in some cases R3C did not improve results and instead led to a decrease in matching rates, particularly for older children in the NITG dataset. For instance, when using FingerGAN, performance dropped from 45.73% to 34.03% at FAR=0.1%, as the recursive process introduced segmentation noise that ultimately degraded TAR by almost 25.6%.

Nevertheless, results indicate that the proposed method can improve matching performance, and its use during the enrollment of newborn fingerprints may help capture ridge–valley structures often missed by standalone enhancement methods. This, in turn, could benefit minutiae extraction, and support more accurate probe matching without compromising real-time applicability.

To enable such applications of R3C, future research may focus on refining the R3C framework through localized recursion strategies, selectively targeting only under-segmented or noisy regions to reduce the risk of over-segmentation. Additionally, an adaptive parameterization strategy, to adjust the $\alpha$ and $\epsilon$ parameters based on ridge width and image resolution, could eliminate the need for manual parameter tuning. Such adjustments could improve R3C's responsiveness to subject-specific traits, including age-related fingerprint variability. In parallel, advances in deep learning-based enhancement techniques and developing more robust acquisition protocols that ensure the enrollment of higher quality fingerprints could help close the performance gap between adult and infant fingerprint recognition.